\documentclass{preprint}

\usepackage{framed,multirow}

\usepackage{amssymb}
\usepackage{latexsym}

\usepackage{url}
\usepackage[table]{xcolor}

\usepackage{hyperref}

\usepackage{tabularx}
\usepackage{graphicx}
\usepackage{booktabs}
\usepackage[para,online,flushleft]{threeparttable}
\usepackage{makecell}
\usepackage{comment}
\usepackage{paralist}
\usepackage{enumitem}
\usepackage{pdflscape}
\usepackage{longtable}
\usepackage{twemojis}
\usepackage[numbers,comma,sort&compress]{natbib}
\usepackage{pdfpages}

\definecolor{newcolor}{rgb}{.8,.349,.1}

\title{CXR-LT 2024: A MICCAI challenge on long-tailed, multi-label, and zero-shot disease classification from chest X-ray}

\author[1,$\dagger$]{Mingquan Lin}
\author[2,$\dagger$]{Gregory Holste}
\author[2]{Song Wang}
\author[3]{Yiliang Zhou}
\author[3]{Yishu Wei}
\author[4]{Imon Banerjee}
\author[5]{Pengyi Chen}
\author[5,6]{Tianjie Dai}
\author[7]{Yuexi Du}
\author[7,8]{Nicha C. Dvornek}
\author[9]{Yuyan Ge}
\author[10]{Zuwei Guo}
\author[11]{Shouhei Hanaoka}
\author[12]{Dongkyun Kim}
\author[13,14]{Pablo Messina}
\author[15]{Yang Lu}
\author[13,14]{Denis Parra}
\author[16]{Donghyun Son}
\author[13]{Álvaro Soto}
\author[4]{Aisha Urooj}
\author[26]{René Vidal}
\author[11]{Yosuke Yamagishi}
\author[17] {Zefan Yang}
\author[15]{Ruichi Zhang}
\author[18]{Yang Zhou}
\author[20,21,22]{Leo Anthony Celi}
\author[23]{Ronald M. Summers}
\author[24]{Zhiyong Lu}
\author[25]{Hao Chen}
\author[19]{Adam Flanders}
\author[3]{George Shih}
\author[2,*]{Zhangyang Wang}
\author[3,*]{Yifan Peng}

\affil[*]{Corresponding author(s). Email(s): \url{atlaswang@utexas.edu}, \url{yip4002@med.cornell.edu}}
\affil[$\dagger$]{These authors contributed equally to this work.}

\begin{document}

\maketitle

\begin{compactenum}
\vspace{-2em}
\item {Department of Surgery, University of Minnesota, Minneapolis, USA}
\item {Department of Electrical and Computer Engineering, The University of Texas at Austin, Austin, USA}
\item {Department of Population Health Sciences, Weill Cornell Medicine, New York, USA}
\item {Department of Radiology, Mayo Clinic, Phoenix, USA}
\item {Cooperative Medianet Innovation Center, Shanghai Jiao Tong University, Shanghai, China}
\item {Shanghai AI Laboratory, Shanghai, China}
\item {Department of Biomedical Engineering, Yale University, New Haven, USA}
\item {Department of Radiology \& Biomedical Imaging, Yale University, New Haven, USA}
\item {Center for Innovation in Data Engineering and Science, Department of Computer and Information Science, University of Pennsylvania, Philadelphia, USA}
\item {School of Electrical, Computer and Energy Engineering, Arizona State University, Tempe, USA}
\item {Department of Radiology, The University of Tokyo Hospital, Tokyo, Japan}
\item {School of Computer Science, Carnegie Mellon University, Pittsburgh, USA}
\item {Pontificia Universidad Católica de Chile, Santiago, Chile}
\item {Millennium Institute for Intelligent Healthcare Engineering (iHEALTH), National Center for Artificial Intelligence (CENIA), Santiago, Chile}
\item {School of Informatics, Xiamen University, Xiamen, China}
\item {Seoul National University, Seoul, South Korea}
\item  {Department of Biomedical Engineering and Center for Biotechnology and Interdisciplinary Studies, Rensselaer Polytechnic Institute, Troy, NY, USA} 
\item {Institute of High Performance Computing, A*STAR, Singapore}
\item {Department of Radiology, Thomas Jefferson University Hospital, Philadelphia, USA}
\item {Laboratory for Computational Physiology, Massachusetts Institute of Technology, Cambridge, USA}
\item {Division of Pulmonary, Critical Care and Sleep Medicine, Beth Israel Deaconess Medical Center, Boston, USA}
\item {Department of Biostatistics, Harvard T.H. Chan School of Public Health, Boston, USA}
\item {Clinical Center, National Institutes of Health, Bethesda, USA}
\item {National Center for Biotechnology Information, National Library of Medicine, Bethesda, USA}
\item {Department of Computer Science and Engineering, Hong Kong University of Science and Technology, Hong Kong, China}
\item {Center for Innovation in Data Engineering and Science, Departments of Electrical and Systems Engineering \& Radiology, University of Pennsylvania, Philadelphia, USA}
\end{compactenum}

\newpage

\begin{abstract}
The CXR-LT series is a community-driven initiative designed to enhance lung disease classification using chest X-rays (CXR). It tackles challenges in open long-tailed lung disease classification and enhances the measurability of state-of-the-art techniques. The first event, CXR-LT 2023, aimed to achieve these goals by providing high-quality benchmark CXR data for model development and conducting comprehensive evaluations to identify ongoing issues impacting lung disease classification performance. 
Building on the success of CXR-LT 2023, the \textbf{CXR-LT 2024} expands the dataset to 377,110 chest X-rays (CXRs) and 45 disease labels, including 19 new rare disease findings. It also introduces a new focus on zero-shot learning to address limitations identified in the previous event. Specifically, CXR-LT 2024 features three tasks: (i) long-tailed classification on a large, noisy test set, (ii) long-tailed classification on a manually annotated "gold standard" subset, and (iii) zero-shot generalization to five previously unseen disease findings.
This paper provides an overview of CXR-LT 2024, detailing the data curation process and consolidating state-of-the-art solutions, including the use of multimodal models for rare disease detection, advanced generative approaches to handle noisy labels, and zero-shot learning strategies for unseen diseases. Additionally, the expanded dataset enhances disease coverage to better represent real-world clinical settings, offering a valuable resource for future research. By synthesizing the insights and innovations of participating teams, we aim to advance the development of clinically realistic and generalizable diagnostic models for chest radiography.
\end{abstract}



\section{Introduction}

The CXR-LT series marks a community-driven initiative to improve lung disease classification using chest X-rays (CXR) that addresses challenges in open long-tailed lung disease classification and advances the measurability of state-of-the-art techniques \citep{holste2022long}. These goals were pursued during the first event, CXR-LT 2023 \citep{holste2024towards}, by offering high-quality benchmark CXR data for model development and conducting detailed evaluations to identify persistent issues affecting lung disease classification performance. CXR-LT 2023 attracted significant attention, with 59 teams yielding over 500 unique submissions. Since then, the task setup and data have provided a foundation for numerous studies \citep{hong2024evolution-aware-l, huijben2024denoising-f, park2024fine-grained-b,li2024improving}.

As the second event in the series, CXR-LT 2024 maintains the general design and goals of its predecessor while introducing a new emphasis on zero-shot learning. This addition addresses a limitation identified in CXR-LT 2023.
The vast number of unique radiological findings, estimated to exceed 4,500 
\citep{budovec2014informatics}, suggests that the actual distribution of clinical findings on CXR is at least two orders of magnitude greater than what current benchmarks can offer.\footnote{\url{http://www.gamuts.net/about.php}} Therefore, effectively addressing the ``long-tail'' of radiological abnormal findings necessitates the development of a model that can generalize to new classes in a ``zero-shot" manner. 

This paper provides an overview of the CXR-LT 2024 challenge, including two long-tailed tasks that attracted extensive participation and one newly introduced zero-shot task. Task 1 and Task 2 focus on long-tailed classification, with Task 1 using a large, noisy test set and Task 2 using a small, manually annotated test set. Task 3 concerns zero-shot generalization to previously unseen diseases. 
Each task adheres to the general framework established by CXR-LT 2023, providing participants with a large, automatically labeled training set consisting of over 250,000 CXR images with 40 binary disease labels. The final submissions from participants are evaluated against a separate held-out test set prepared in a similar manner. 

In the following sections, we introduce each task setting and outline the evaluation criteria. Next, we detail the data curation process before presenting the results for each task. We then consolidate key insights from top-performing solutions and provide practical perspectives. Finally, we use our findings to suggest a path forward for few- and zero-shot disease classification, emphasizing the potential of leveraging multimodal foundation models.

\section{Methods}

\subsection{Main tasks}

The CXR-LT 2024 challenge includes three tasks: 1) long-tailed classification on a large, noisy test set, (2) long-tailed classification on a small, manually annotated test set, and (3) zero-shot generalization to previously unseen diseases. All can be formulated as multi-label classification problems. 

Given the severe label imbalance in these tasks, the primary evaluation metric was mean average precision (mAP), specifically the ``macro-averaged" AP across classes. While the area under the receiver operating characteristic curve (AUROC) is often used for similar datasets \citep{wang2017chestx,seyyed2020chexclusion}, it can be heavily inflated in the presence of class imbalance \citep{fernandez2018learning,davis2006relationship}. In contrast, mAP is more suitable for long-tailed, multi-label settings as it measures the performance across decision thresholds without degrading under-class imbalance \citep{rethmeier2022long}. For thoroughness, mean AUROC (mAUROC) and mean F1 score (mF1) -- with a threshold of 0.5 -- were computed as auxiliary classification metrics. We also calculated the mean expected calibration error (ECE) \citep{naeini2015obtaining} to quantify bias. To further enhance clinical interpretability, we also report per-class F1 scores, as well as macro- and micro-averaged F1 scores and false-negative rates for critical findings, in addition to the challenge's primary evaluation metric. We believe these additions provide a more granular understanding of model performance in practical settings.

\subsection{Dataset curation}

Table \ref{tab:dataset} lists the characteristics of the datasets used in the three tasks. The same training dataset is utilized for all three tasks. Similarly, the same development dataset is shared across all three tasks, with Task 3 focusing on five unseen classes. Task 1 and Task 3 share the same test set; however, Task 3 explicitly evaluates performance on the five unseen classes. Task 2 is a subset of Task 1, containing 26 manually annotated classes. Fig.~\ref{fig1} lists the 45 classes, with the five unseen classes being Bulla, Cardiomyopathy, Hilum, Osteopenia, and Scoliosis. The remaining 40 classes exclude these five unseen classes, while the 14 classes are derived from the original MIMIC-CXR dataset and the 12 additional classes introduced in CXR-LT 2023. Fig.~\ref{figexample} shows example chest X-rays from the challenge dataset, where each image contains multiple annotated abnormalities.

\begin{table}
    \caption{\label{tab:dataset} Characteristics of the datasets used in the three tasks.}
    \centering
    \small
    \begin{tabular}{lrrcrrcrr}
    \toprule
    & \multicolumn{2}{c}{Task 1} && \multicolumn{2}{c}{Task 2} && \multicolumn{2}{c}{Task 3} \\ 
    \cmidrule(rl){2-3}\cmidrule(rl){5-6}\cmidrule(rl){8-9}
    Dataset & Samples & Labels && Samples & Labels && Samples & Labels\\\midrule
    Train & 258,871 & 40 && 258,871 & 40 && 258,871 & 40 \\ 
    Development & 39,293 & 40 && 39,293 & 40 && 39,293 & 5 \\ 
    Test & 78,946 & 40 && 406 & 26 && 78,946 & 5 \\ 
    \bottomrule
    \end{tabular}
\end{table}

In this section, we detail the data curation process of two datasets: (i) the automatically labeled CXR-LT dataset used in Tasks 1 and 3, and (ii) a manually annotated ``gold standard" test set used in Task 2.

\subsubsection{CXR-LT dataset}
\label{sec:data}

The CXR-LT challenge dataset was developed by expanding the label set of the MIMIC-CXR dataset \citep{johnson2019mimic},\footnote{\url{https://physionet.org/content/mimic-cxr/2.0.0/}} resulting in a more complex, long-tailed label distribution. 
This year, newly added clinical findings were selected from sources including the disease list of the PadChest dataset \citep{bustos2020padchest} and the Fleischner glossary of thoracic imaging terms \citep{hansell2008fleischner}. After ensuring that a sufficient number of occurrences were observed in the dataset for reliable evaluation, these \textit{19 new disease findings} are:
\begin{inparaenum}[(1)]
    \item Adenopathy,
    \item Azygos Lobe,
    \item Clavicle Fracture,
    \item Fissure,
    \item Hydropneumothorax,
    \item Infarction,
    \item Kyphosis,
    \item Lobar Atelectasis,
    \item Pleural Other,
    \item Pulmonary Embolism,
    \item Pulmonary Hypertension,
    \item Rib Fracture,
    \item Round Atelectasis,
    \item Tuberculosis,
    \item Bulla,
    \item Cardiomyopathy,
    \item Hilum,
    \item Osteopenia,
    \item Scoliosis.
\end{inparaenum} 
The last five abnormal findings -- Bulla, Cardiomyopathy, Hilum, Osteopenia, and Scoliosis -- were not included in the challenge training set and were held out for zero-shot evaluation in Task 3.

In addition, we replaced the ``No Finding" class with a more intuitive ``Normal" class. ``No Finding" indicated that none of the abnormal findings in the label set were present. For instance, with the original 14 MIMIC-CXR classes, ``No Finding" meant that none of these 14 findings were present; however, when expanding the label set to 26 classes, this ``No Finding" label may, in fact, include one of the 12 added abnormalities. To avoid unclear interpretation of this label across tasks, we curated new labels for a simplified ``Normal" class, signifying that no cardiopulmonary disease or abnormality was found in the report.
Like in 2023, the radiology reports for each CXR study were parsed using RadText \citep{wang2022radtext}, a radiology text analysis tool, to extract the presence status of new diseases. 

\begin{figure*}
    \centering
    \includegraphics[width=\textwidth]{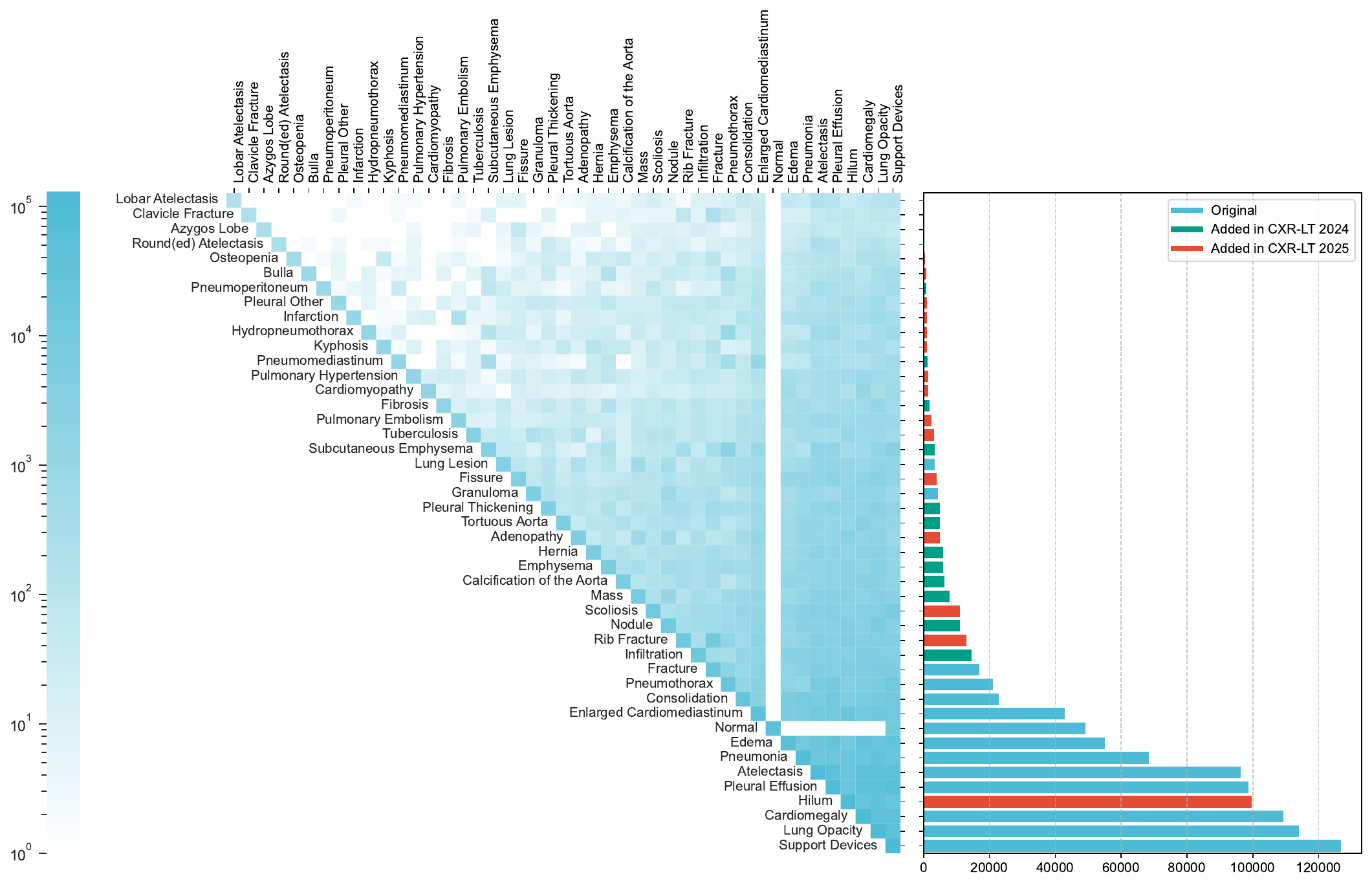}
    \caption{Long-tailed distribution of the CXR-LT 2024 challenge dataset. The dataset was formed by extending the MIMIC-CXR benchmark to include 12 new clinical findings (red) by parsing radiology reports.}
    \label{fig1}
\end{figure*}

\begin{figure*}
    \centering
    \includegraphics[width=\textwidth]{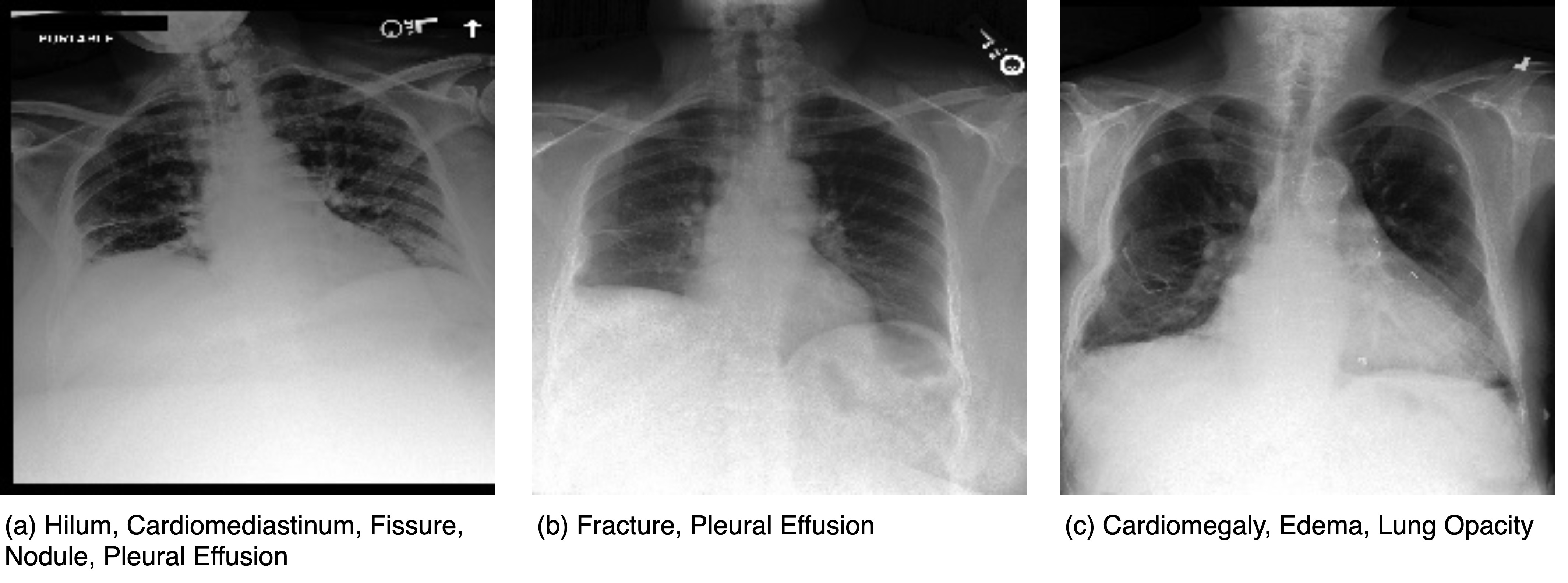}
    \caption{Representative chest X-rays from the challenge dataset, each demonstrating multiple findings. (a) Includes the Hilum label (new in CXR-LT 2024); (b) shows Fracture (introduced in CXR-LT 2023); and (c) displays original MIMIC-CXR labels (Cardiomegaly, Edema, Lung Opacity).}
    \label{figexample}
\end{figure*}

The final dataset included 377,110 CXR images, each labeled with one or more of the 45 diseases, following a long-tailed distribution (Fig.~\ref{fig1}). Like CXR-LT 2023, we opted to use image data from the MIMIC-CXR-JPG dataset \citep{johnson2019mimicjpg} to increase accessibility and reduce the burden of storing this dataset ($\sim$600 GB vs. $\sim$4.7 TB using the raw DICOM data provided by MIMIC-CXR).\footnote{\url{https://physionet.org/content/mimic-cxr-jpg/2.0.0/}} The dataset was randomly partitioned into training (70\%), development (10\%), and test (20\%) sets at the patient level; critically, this split was unique to CXR-LT 2024, meaning participants could not re-use models from the previous year's challenge. Participants had access to all images, but were provided with labels only for the training set.

\subsubsection{Gold standard test set}
\label{sec:gold-data}

In our overview of CXR-LT 2023 \citep{holste2024towards}, we used a manually annotated ``gold standard" set derived from the challenge test set to evaluate the differences in manual vs. automated annotation as well as how top-performing solutions fared on this test set with reduced label noise. Specifically, six annotators reviewed 406 MIMIC-CXR radiology reports for the presence or absence of 26 disease findings considered in CXR-LT 2023. For complete data curation details of this gold standard set, please see \citet{holste2024towards}. This dataset provides a high-quality benchmark for evaluating model performance on a smaller, manually vetted test set. This year, in CXR-LT 2024, we used this gold standard dataset as the test set in Task 2.

\subsection{Schedule}

Table \ref{tab:schedule} shows the task schedule. The challenge was conducted on the CodaLab platform,\footnote{\url{https://codalab.lisn.upsaclay.fr/competitions/18601}, \url{https://codalab.lisn.upsaclay.fr/competitions/18603}, \url{https://codalab.lisn.upsaclay.fr/competitions/18604}} with a separate CodaLab page for each of the three tasks \citep{codalab_competitions_JMLR}. Any registered CodaLab user could apply, but would only be accepted after submitting proof of the necessary PhysioNet credentials required to access MIMIC-CXR-JPG.

During the Development Phase (May 1, 2024 - August 26, 2024), registered participants downloaded the labeled training set and the unlabeled development set, from which they generated a comma-separated values (CSV) file with predictions to upload. Submissions were evaluated on the held-out development set, and results were updated on a live, public leaderboard. 
During the Test Phase (August 26, 2024 - September 6, 2024), test set images (without labels) were released. Participants were asked to submit CSV files with test set predictions for final evaluation and ranking in each task. The leaderboard was hidden in this phase, and each team's best-scoring submission was retained. The final Test Phase leaderboard ranked participants primarily by mAP, then by mAUROC in the event of ties. 

\begin{table}
\caption{Schedule of CXR-LT 2024.}
\label{tab:schedule}
\small
\centering
\begin{tabular}{lrr}
\toprule
\textbf{Event} & \textbf{Date} & \textbf{Teams}\\
\midrule
Registration & May 1, 2024 & 61\\
\arrayrulecolor{black!30}\midrule
Development phase &  May 1, 2024 & 29\\
\hspace{2em}Training data\\
\hspace{2em}Development data\\
\hspace{2em}Leaderboard\\
\midrule
Test phase & & 17\\
\hspace{2em}Test data & Aug 26, 2024 &  \\
\hspace{2em}Submission & Sept 6, 2024\\
\midrule
Workshop &  Oct 10, 2024 & 9\\
\arrayrulecolor{black}\bottomrule
\end{tabular}
\end{table}

\section{Results}

\subsection{Participation}

The CXR-LT challenge received 96 team applications on CodaLab, of which 61 were approved after providing proof of credentialed access to MIMIC-CXR-JPG \citep{johnson2019mimicjpg}. During the Development Phase, 29 teams participated, submitting a total of 661, 349, and 364 unique submissions to the public leaderboard for Tasks 1, 2, and 3, respectively. In the final Test Phase, a total of 17 teams participated.
We selected the top 9 teams for the invitation to present at the CXR-LT 2024 challenge event at MICCAI 2024 \footnote{\url{https://cxr-lt.github.io/CXR-LT-2024/}} and for inclusion in this study. Since two teams excelled in both Tasks 1 and 2, this comprised the top 4 solutions in Tasks 1 and 2 as well as the top 3 solutions for the zero-shot Task 3.
Table \ref{tab:overview} summarizes the top-performing groups participating in one or more of these tasks and system descriptions. Additional details, including all presentation slides, are available on GitHub\footnote{\url{https://github.com/CXR-LT/CXR-LT-2024}}, allowing readers to explore the specifics of all methods in greater depth.

\begin{table}[t]
    \caption{\label{tab:overview}Overview of top-performing CXR-LT 2024 challenge solutions. ENS - ensemble; LRW - loss reweighting; VL - vision-language.}
\footnotesize
\centering
\begin{tabular}{l>{\raggedright\arraybackslash}m{2.3cm}llcccl}
\toprule
Team  & Institution & \makecell[l]{Image\\Resolution} & Backbone & ENS & LRW & VL & Pre-training \\
\midrule
A & Arizona State University & 224, 384 & \makecell[l]{ConvNeXt-S\\ConvNeXt-B\\ConvNeXt-T\\ConvNeXt V2-B} & \checkmark & & \checkmark & ImageNet \\
\arrayrulecolor{black!30}\midrule
B & Shanghai Jiaotong University& 512 & EfficientNetV2-L & \checkmark & \checkmark & \checkmark & \makecell[l]{ImageNet $\rightarrow$ NIH, CheXpert,\\ VinDr-CXR, BRAX} \\
\midrule
C & Yale University& 1024 & \makecell[l]{ConvNeXt-S\\EfficientNetV2-S} & \checkmark & \checkmark & \checkmark & ImageNet $\rightarrow$ MIMIC-CXR \\
\midrule
D & Carnegie Mellon University  & 1024 & ConvNeXt-S & & \checkmark & \checkmark & \makecell[l]{ImageNet $\rightarrow$ CheXpert,\\ NIH, VinDr-CXR} \\
\midrule
E & Rensselaer Polytechnic Institute & \makecell[l]{336, 448,\\512} & ViT-L & \checkmark & & & \makecell[l]{MIMIC-CXR, CheXpert, \\PadChest, NIH, BRAX} \\
\midrule
F & The University of Tokyo & 384, 512 & \makecell[l]{ConvNeXt V2-S\\MaxViT-T} & \checkmark & \checkmark & & ImageNet $\rightarrow$ NIH \\
\midrule
G & University of Pennsylvania & 224 & ViT-L & \checkmark & \checkmark & \checkmark & \makecell[l]{ImageNet $\rightarrow$ MIMIC-CXR,\\CXR-Concepts,\\Chest ImaGenome, CXR-LT} \\
\midrule
H & Xiamen University & 224 & ResNet50 & \checkmark & & \checkmark & None \\
\midrule
I & Pontifical Catholic University of Chile & 384, 416 & \makecell[l]{DenseNet121\\SigLIP Base\\ConvNeXt-S\\Uniformer} & \checkmark & \checkmark & \checkmark & \makecell[l]{ImageNet $\rightarrow$ MIMIC-CXR, IU X-ray,\\Chest ImaGenome, CheXpert,\\CheXlocalize, VinDr-CXR} \\
\arrayrulecolor{black!}
\bottomrule
\end{tabular}
\end{table}

\subsection{System descriptions}




\paragraph{Team A: \texttt{zguo}}

This team proposed the ChexFusion+ model for CXR-LT classification, participating in Task 1 and Task 2. Their approach leveraged ensembles of 12 multi-resolution ConvNeXt \citep{liu2022convnet} models trained with synthetically generated long-tail data. Specifically, they used input prompts and random Gaussian noise to generate two images through a conditional denoising U-Net and a variational autoencoder decoder (VAE) decoder. The two generated images, together with the input prompts and the corresponding MIMIC-CXR image, were used as input for the ConvNeXt for training. To address the data imbalance problem in the tail cases, they used pre-trained large generative models to generate 100 new images per rare disease class. These synthetic images were generated using carefully constructed prompts that specified multiple co-occurring thoracic abnormalities, such as ``Round(ed) Atelectasis, Pneumothorax, Pleural Effusion, Lung Opacity, and Atelectasis," to reflect realistic radiographic comorbidities.


\paragraph{Team B: \texttt{tianjie\_dai}}

This team leveraged a multimodal ensemble approach to address the imbalanced, multi-label classification challenge in the CXR-LT tasks. Specifically, they employed an ensemble of EfficientNetV2-Large \citep{tan2021efficientnetv2} and PubMedBERT \citep{pubmedbert} models, fine-tuned on a Unified Language Medical System (UMLS) knowledge graph \citep{dai2024unichest}, to integrate both image and text features. To address class imbalance, they used an asymmetric loss function \citep{kim2023chexfusion}, assigning higher weights to rare classes. Test-time augmentation techniques were applied to improve model robustness and generalization, including resizing, cropping, and flipping. Additionally, they incorporated external datasets from multiple sources, such as ChestXRay-14 \citep{wang2017chestx}, CheXpert \citep{irvin2019chexpert}, VinDr-CXR \citep{nguyen2022vindr}, and BRAX \citep{reis2022brax}, to enhance representation learning of rare disease labels.

\paragraph{Team C: \texttt{XYPB}}

 This team addressed the CXR-LT challenge with a multimodal ensemble approach, leveraging multi-view and multi-scale image alignment to enhance classification in the long-tailed, multi-label setting. Their method builds on the CLEFT \citep{du2024cleft} and MaMA \citep{du2024multi} frameworks, incorporating contrastive language-image pretraining (CLIP) with additional image-to-image and image-to-text contrastive learning across multi-view image pairs from chest X-ray studies. This approach allowed their model to capture information from varied perspectives, such as postero-anterior and lateral views. To tackle the multi-scale nature of medical imaging, they introduced a Symmetric Local Cross-Attention (SLA) alignment module that models region-specific visual-text correlations through cross-attention, aligning local image patches with descriptive text segments. To counteract class imbalance, they used a weighted asymmetric loss \citep{ridnik2021asymmetric}. For the image encoder, they utilized ConvNeXt-S \citep{liu2022convnet} and EfficientNet V2-S \citep{tan2021efficientnetv2} backbones, pre-trained on ImageNet for classification and MIMIC-CXR using a CLIP-based approach, for Task 1 and Task 2, and use a parameter efficient fine-tuned medical LLM, \textit{i.e.}, BioMedLM \citep{bolton2024biomedlm}, as their language encoder.

\paragraph{Team D: \texttt{dongkyunk}}
This team implemented a two-stage framework designed to effectively leverage the multiple views available for each patient. In the first stage, a single model was trained using the ML-Decoder \citep{ridnik2023ml} classification head alongside Noisy Student \citep{xie2020self} self-training. In the second stage, a Transformer-based model called CheXFusion was introduced to aggregate multi-view features \citep{kim2023chexfusion}. The feature aggregation in CheXFusion is analogous to encoding multiple sentences in natural language processing, where each sentence represents a single chest X-ray image. Additionally, a weighted version of the asymmetric loss \citep{ridnik2021asymmetric} was employed to address inter-class imbalance from the long-tailed distribution of diseases and intra-class imbalance due to the predominance of negative labels in multi-label classification.

\paragraph{Team E: \texttt{yangz16}}
This team tackled the CXR-LT challenge with a foundation model-based approach. Their methodology incorporated three key components: DINOv2 foundation models \citep{oquab2023dinov2} utilizing the ViT-Large network architecture \citep{neil2020transformers} as the backbone, the ML-Decoder \citep{neil2020transformers} classification head, and multi-view/multi-resolution ensembling. The DINOv2 models were pre-trained on over 710,000 chest X-rays from diverse datasets, including MIMIC-CXR \citep{johnson2019mimic}, CheXpert \citep{irvin2019chexpert}, PadChest \citep{bustos2020padchest}, NIH Chest X-ray14 \citep{wang2017chestx}, and BRAX \citep{reis2022brax}, using self-supervised learning that combined a self-distillation loss and masked image modeling to learn robust representations. The ML-Decoder mapped local features from the foundation model to disease-specific predictions for classification, employing attention mechanisms to achieve finer localization of disease findings.

\paragraph{Team F: \texttt{YYama}}

This team utilized an ensemble of ConvNeXt V2 \citep{liu2022convnet} and MaxViT \citep{tu2022maxvit} models with domain-specific pretraining and view-based aggregation. The ConvNeXt V2 models were pre-trained on ImageNet, while the MaxViT model was further pretrained on the NIH Chest X-ray dataset. To address class imbalance, they applied an asymmetric loss function \citep{ridnik2021asymmetric} combined with class weights, assigning higher importance to rare classes. Additionally, they implemented a view-based prediction aggregation method, combining predictions from frontal and lateral views, with a weighted average favoring the frontal view.

\paragraph{Team G: \texttt{yyge}}

This team employed a dual-model strategy combining Vision-Language Models (VLM) and Multi-View Vision Models (MVM) for zero-shot and multi-label disease classification in chest X-rays. The VLM integrated DINOv2 \citep{oquab2023dinov2} as the image encoder and BERT \citep{boecking2022making} for text encoding, utilizing fine-grained disease descriptions from ChatGPT \citep{achiam2023gpt}. The model was first pretrained on domain-specific datasets, and then was fine-tuned on the CXR-LT training set with the weighted binary cross-entropy loss for class imbalance. Meanwhile, the MVM converted zero-shot tasks into few-shot problems by mining disease-specific examples from radiology reports and aggregated multi-view features using DINOv2 and lightweight Transformers. Using a weighted asymmetric loss \citep{kim2023chexfusion} and multi-view learning further addressed long-tail distributions. This combined framework effectively captured domain-specific knowledge and balanced performance across seen and unseen diseases.

\paragraph{Team H: \texttt{ZhangRuichi}}

This team utilized a Visual-Language Model (VLM) inspired by MedKLIP \citep{wu2023medklip}, incorporating anatomical and textual information to enhance generalization and performance. The authors used a ResNet50 \citep{he2016deep} architecture without pretraining for image encoding, as pretraining on ImageNet may introduce biases due to domain and distribution differences between natural images and chest X-ray datasets. By training the model from scratch, the authors aim to mitigate these domain and distribution gaps and better tailor the model to the characteristics of chest X-ray images. To enable zero-shot capability, categorical labels were augmented with GPT-4-generated descriptions, and BioClinical BERT \citep{alsentzer2019publicly} was employed for text encoding, capturing rich semantic information. Anatomical location data from CheXpert was integrated, mapping MIMIC-CXR diseases to corresponding regions to address the lack of fine-grained labels. For training, they applied a cross-entropy loss to improve accuracy and a contrastive loss to link diseases with anatomical regions. During testing, a class-wise ensemble strategy and test-time fusion of predictions from different patient views were implemented to improve overall accuracy.

\paragraph{Team I: \texttt{pamessina}}

This team developed a multimodal model for the zero-shot classification task. The text encoder is the CXR Fact Encoder (CXRFE) \citep{messina2024extracting}, which computes fact embeddings from short factual sentences. The text encoder remains frozen during training. The image encoder is trained end-to-end, and the team experimented with several architectures, including DenseNet121 \citep{huang2017densely}, SigLIP Base \citep{zhai2023sigmoid}, ConvNext-Small \citep{liu2022convnet}, and Uniformer \citep{li2022uniformer}. The model uses the fact embeddings to modulate local and global features from the image encoder via FiLM \citep{perez2018film} to predict both a binary coarse segmentation mask (representing the visual grounding of the fact) and a global binary classification of the fact for the entire image. The best results were achieved using an ensemble of 21 models, each with a different configuration of image encoder and training data. This work also used GPT-4 \citep{achiam2023gpt} as an automatic labeler for MIMIC-CXR reports and included additional datasets, some of which contained bounding boxes to provide visual grounding supervision.

\subsection{Task 1 primary evaluation results}

\paragraph{CXR-LT Test Phase results}

The results of the top 4 teams in Task 1 are listed in Table \ref{tab:results1}. Team A took first place with an mAP of 0.281, Team B came in second with an mAP of 0.279, while Teams C and D both achieved an mAP of 0.277. However, Team C ranked third due to a higher mAUC. In CXR-LT 2023, the top 4 teams achieved mAP scores ranging from 0.349 to 0.372, much higher than this year's scores due to the inclusion of 19 new rare classes in CXR-LT 2024.
When evaluating this year's top solutions on the set of 26 CXR-LT 2023 labels, we observe mAP scores of 0.371, 0.373, 0.371, and 0.370, respectively. Compared to CXR-LT 2023, top performers in Task 1 improved their overall performance in these classes (e.g., the second-place finisher in CXR-LT 2023 reached 0.354 mAP). Supplementary Table 1 details the performance of each class for the top four teams. Additionally, Supplementary Tables 2 and 3 present per-class F1 scores, macro- and micro-averaged F1 scores, and false-negative rates for critical findings.

\begin{table}
    \caption{\label{tab:results1}mAP of top-4 team's final model on all 40 classes evaluated on the test set in Task 1. mAUROC, mF1, and mECE are also presented, with the numbers in parentheses indicating the rankings based on the corresponding evaluation metric.}
    \centering
    \small
    \begin{tabular}{ccrr@{~~}ccr@{~~}ccr@{~~}c}
    \toprule
     &&&\multicolumn{2}{c}{mAUROC} && \multicolumn{2}{c}{mF1} && \multicolumn{2}{c}{mECE}\\  
     \cmidrule{4-5}\cmidrule{7-8}\cmidrule{10-11}
     Ranking & Team & mAP & \hspace{7ex} & \twemoji{trophy} &&& 
     \twemoji{trophy} &&& \twemoji{trophy} \\\midrule
    1 & A & 0.281 & 0.847 & (3) &&0.289 & (3) && 0.589 & (4)\\
    2 & B & 0.279 & 0.843 & (5) &&0.286 & (4) && 0.592 & (6) \\
    3 & C & 0.277 & 0.849 & (1) && 0.299 & (1) && 0.603 & (8) \\
    4 & D & 0.277 & 0.842 & (6)	&& 0.285 & (5) && 0.602 & (7) \\
   \bottomrule
    \end{tabular}
\end{table}

\paragraph{Long-tailed classification performance}

To examine predictive performance by label frequency, we split the 40 target classes into ``head'' ($>$10\%), ``medium'' (1\%--10\%), and ``tail'' ($<$1\%) categories based on their prevalence in the training set, consisting of 9, 14, and 16 categories, respectively. Category-wise mAP is presented in Table \ref{tab:lt-results}, as well as a ``category-wise average'' of head, medium, and tail mAP. Team \texttt{A} not only achieved the highest overall performance but also excelled in the ``tail'' group. However, the top three performances in the ``tail'' group were very close. Team \texttt{A} used pretrained large generative models to generate new images for these tail cases, while Teams \texttt{B} and \texttt{C} applied loss reweighting, indicating that both approaches can improve performance in the ``tail'' group.

\begin{table}
    \caption{Long-tailed classification performance on ``head", ``medium", and ``tail" classes by average mAP within each category. These categories were determined by the relative frequency of each class in the training set (denoted in parentheses). The rightmost column denotes the average of head, medium, and tail mAP. The best mAP in each column appears in bold.}
    \label{tab:lt-results}
    \small
    \centering
    \begin{tabular}{cccccc}
        \toprule
         Team & Overall & Head & Medium & Tail & Avg \\ 
         & & ($>$10\%) & (1-10\%) & ($<$1\%)\\\midrule
        A & \textbf{0.281} & 0.567 & 0.263 & \textbf{0.136} & \textbf{0.322}  \\
        B & 0.279 & 0.569 & 0.260 & 0.133 & 0.321  \\
        C & 0.277 & \textbf{0.570} & 0.253 & \textbf{0.136} & 0.320  \\
        D & 0.277 & 0.568 & \textbf{0.264} & 0.125 & 0.320 \\
 \bottomrule
    \end{tabular}
\end{table}

\subsection{Task 2 primary evaluation results}

The results of the top 4 teams in Task 2 are listed in Table \ref{tab:results2}. The first place went to Team \texttt{C} with an mAP of 0.526. Team \texttt{E} placed second with an mAP of 0.511, Team \texttt{A} secured third with an mAP of 0.511, and Team \texttt{F} placed fourth with an mAP of 0.509. All four teams used ensembling methods to improve model performance, achieving results similar to last year; for instance, top CXR-LT 2023 performers achieved mAP scores of 0.519, 0.518, and 0.519 on the same gold standard test set. Supplementary Table 4 provides the detailed class-specific performance for these top teams. Additionally, Supplementary Tables 5 and 6 present per-class F1 scores, macro- and micro-averaged F1 scores, and false-negative rates for critical findings.

\begin{figure}
    \centering
    \includegraphics[scale=0.42]{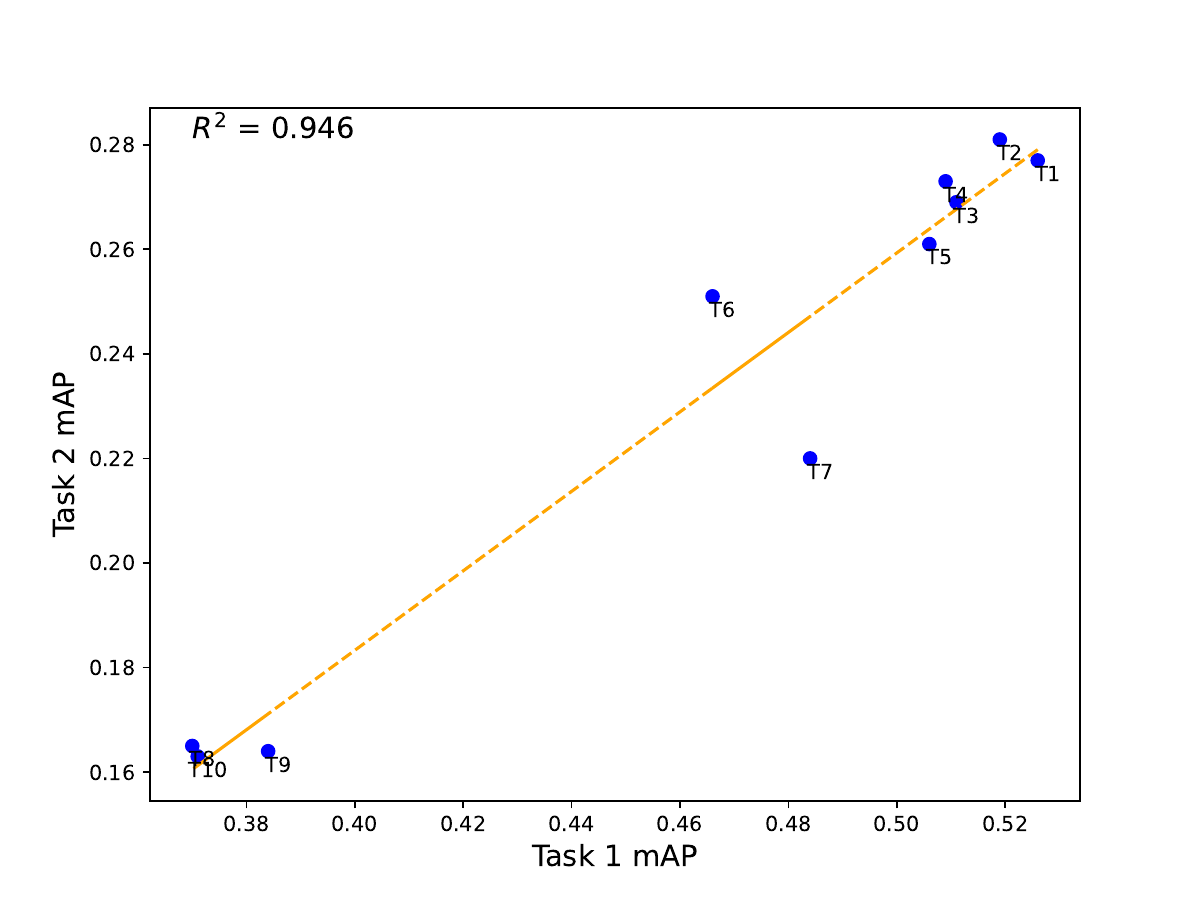}
    \caption{Comparison of performance on CXR-LT Task 1 data (Section \ref{sec:data}) and gold standard Task 2 data (Section \ref{sec:gold-data}).}
    \label{fig:test-vs-gold}
\end{figure}

As mentioned in Section \ref{sec:gold-data}, the test set in Task 2 is the subset of the test set in Task 1, with only 26 manually annotated labels. Table \ref{tab:results1} and \ref{tab:results2} show that the first-place team in Task 2 ranked third in Task 1, while the second-place team in Task 2 was first in Task 1. Additionally, from the challenge leaderboards\footnote{\url{https://codalab.lisn.upsaclay.fr/competitions/18603\#results}} \footnote{\url{https://codalab.lisn.upsaclay.fr/competitions/18601\#results}}, we can see that the third- and fourth-place teams in Task 2 ranked sixth and fifth in Task 1, with mAP scores of 0.269 and 0.273, respectively. We selected ten teams that submitted their results for both tasks, named them T1 to T10, and analyzed their performance based on the results. Despite a large distribution shift between these datasets, the overall performance consistency remained stable between Tasks 1 and 2
(Fig.~\ref{fig:test-vs-gold}; $R^2 = 0.946$, $r=0.972$).

\begin{table}
    \caption{\label{tab:results2}mAP of top-4 team's final model on all 26 classes evaluated on the Gold standard test set in Task 2. mAUROC, mF1, and mECE are also presented, with the numbers in parentheses indicating the rankings based on the corresponding evaluation metric.}
    \centering
    \small
    \begin{tabular}{ccrr@{~~}ccr@{~~}ccr@{~~}c}
    \toprule
     &&&\multicolumn{2}{c}{mAUROC} && \multicolumn{2}{c}{mF1} && \multicolumn{2}{c}{mECE}\\  
     \cmidrule{4-5}\cmidrule{7-8}\cmidrule{10-11}
     Ranking & Team & mAP & \hspace{7ex} & \twemoji{trophy} &&& \twemoji{trophy} &&& \twemoji{trophy} \\\midrule
    1 & C & 0.526 & 0.833 & (3) && 0.499 & (1)	&& 0.464 & (6) \\
    2  & A & 0.519 & 0.834 & (2) && 0.471 & (4) && 0.457 & (30 \\
    3 & E & 0.511 & 0.836 & (1) &&0.265 & (9) && 0.744 & (10) \\
    4 & F & 0.509 & 0.829 & (5) && 0.474 & (3) && 0.462 & (5)\\
   \bottomrule
    \end{tabular}
\end{table}

\subsection{Task 3 primary evaluation results}


Table \ref{tab:results3} presents results for the top 3 performing teams in Task 3. Team \texttt{G} secured the first place with an mAP of 0.129. Following closely, Team \texttt{H} earned second place with an mAP of 0.116, while Team \texttt{I} came in third with an mAP of 0.110. Compared to the other tasks, the relatively low performance in Task 3 can be attributed to the challenging zero-shot nature of detecting findings that were never seen during training. Supplementary Table 7 details the class-specific performance of these top teams. Additionally, Supplementary Tables 8 and 9 report per-class F1 scores, macro- and micro-averaged F1 scores, and false-negative rates for critical findings.

\begin{table}[!t]
    \caption{\label{tab:results3}Performance evaluation of the final models from the top 3 teams on the test set for all five unseen classes in Task 3. mAUROC, mF1, and mECE are also presented, with the numbers in parentheses indicating the rankings based on the corresponding evaluation metric.}
    \centering
    \small
    \begin{tabular}{ccrr@{~~}ccr@{~~}ccr@{~~}c}
    \toprule
     &&&\multicolumn{2}{c}{mAUROC} && \multicolumn{2}{c}{mF1} && \multicolumn{2}{c}{mECE}\\  
     \cmidrule{4-5}\cmidrule{7-8}\cmidrule{10-11}
     Ranking & Team & mAP & \hspace{7ex} & \twemoji{trophy} &&& \twemoji{trophy} &&& \twemoji{trophy} \\\midrule
    1 & G  & 0.129 & 0.741 & (2) && 0.075 & (6) && 0.817 & (8)\\
    2  & H & 0.116 & 0.673 & (8) && 0.035 & (7) && 0.907 & (9) \\
    3 & I & 0.110 & 0.744 & (1) && 0.094 & (4) && 0.711 & (6)\\
   \bottomrule
    \end{tabular}
\end{table}

\subsection{Comparison of rule-based and GPT-4o labeling}

In the CXR-LT dataset, labels were initially generated using a rule-based method. Approaches like this have proven successful in automatically labeling existing CXR datasets \citep{irvin2019chexpert,wang2017chestx,bustos2020padchest}, but recent developments in large language models (LLMs) suggest they may be useful candidates for this task. Team \texttt{I} opted to use GPT-4 to label the data from MIMIC-CXR reports, instead of relying on the rule-based labels. With a dataset of 406 manually annotated samples, we calculated the precision to assess whether large language models could generate more accurate labels. Table \ref{tab:comp-results} lists the performance comparison between the rule-based method and GPT-4o, using prompts suggested by \citet{wei2024enhancing}. The rule-based method achieved a precision of 0.711, whereas GPT-4o reached a precision of 0.786. 

\begin{table}[t]
    \caption{\label{tab:comp-results}Performance of models on long-tailed, multi-label disease classification evaluated using micro-precision on our gold standard test set.}
    \small
    \centering
    \begin{tabular}{@{}l@{~~}ccr@{}}
    \toprule
     & Rule-based & GPT-4o \\ \midrule
    Atelectasis & 0.611 & 0.590 \\
    Calcification of the Aorta & 1.000 & 0.857  \\
    Cardiomegaly & 0.769 & 0.938 \\
    Consolidation & 0.816 & 0.849 \\
    Edema & 0.638 & 0.804  \\
    Emphysema & 0.609 & 0.639  \\
    Enlarged Cardiomediastinum & 0.583 & 1.000  \\
    Fibrosis & 0.667 & 0.682 \\
    Fracture & 0.870 & 0.937 \\
    Hernia & 0.633 & 0.810  \\
    Infiltration & 0.261 & 0.889  \\
    Lung Lesion & 0.161 & 0.000  \\
    Lung Opacity & 0.853 & 0.984  \\
    Mass & 0.513 & 0.810  \\
    Normal & 0.917 & 0.972 \\
    Nodule & 0.821 & 0.844  \\
    Pleural Effusion & 0.798 & 0.812 \\
    Pleural Other & 0.810 & 0.500 \\
    Pleural Thickening & 1.000 & 0.815 \\
    Pneumomediastinum & 0.875 & 0.889 \\
    Pneumonia & 0.191 & 0.435  \\
    Pneumoperitoneum & 0.676 & 0.840  \\
    Pneumothorax & 0.563 & 0.865  \\
    Subcutaneous Emphysema & 0.955 & 0.889  \\
    Support Devices & 0.948 & 0.933  \\
    Tortuous Aorta & 0.958 & 0.861 \\
    \midrule
    Mean & 0.711 & \textbf{0.786}  \\ \bottomrule
    \end{tabular}
\end{table}

\section{Discussion}

\subsection{Themes of top CXR-LT 2024 solutions}

As outlined in Table \ref{tab:overview} the system descriptions, we observe several common themes among top-performing solutions across the three tasks, as well as some unique perspectives. 

\paragraph{Modern Convolutional Neural Network Architectures}

The top-performing solutions commonly used convolutional neural networks (CNNs) as image encoders, continuing the trend from CXR-LT 2023. ConvNeXt emerged as the most popular choice  \citep{liu2022convnet}, followed by EfficientNet \citep{tan2021efficientnetv2}, ResNet \citep{he2016deep}, and DenseNet \citep{huang2017densely}. 
ConvNeXt consistently outperformed other architectures, both in 2023 and 2024. We attribute ConvNeXt's popularity and strong performance to two main factors: (1) its adoption by last year's top-performing solutions, which sets a standard, and (2) its design for scalable performance across different image resolutions, making it well-suited to capture multi-scale information.

\paragraph{Vision Transformers}

While none of the top-performing solutions used Vision Transformers (ViTs) \citep{khan2022transformers} as image encoders in 2023, there has been a noticeable shift in 2024, with four teams adopting ViT-based models. Specifically, Teams \texttt{E} and \texttt{G} dedicated their image encoding entirely to ViTs. In contrast, Teams \texttt{F} and \texttt{I} opted for a hybrid approach, combining ViT-based Transformers with CNNs. We attribute the increased adoption of ViT-based transformers to two main factors: (1) the complementary strengths offered by integrating both CNNs and ViTs \citep{pantelaios2024hybrid}, which can enhance feature extraction and representation capabilities, and (2) the strategic use of additional datasets for pretraining the image encoders, which is particularly beneficial for effectively training ViT-based transformers, improving their robustness and generalization.

\paragraph{Large-Scale Pretraining}

Eight of the nine top-performing solutions relied on supervised pretraining or transfer learning. While some teams used standard ImageNet-pretrained models, several others performed additional pretraining on publicly available ``in-domain" CXR datasets such as ChestXRay-14 \citep{wang2017chestx}, CheXpert \citep{irvin2019chexpert}, VinDr-CXR \citep{nguyen2022vindr}, and BRAX \citep{reis2022brax}. Notably, Teams \texttt{B}, \texttt{C}, \texttt{D}, \texttt{F}, and \texttt{G} employed a multi-stage pretraining strategy, starting with general pretraining on natural images, followed by domain-specific pretraining on CXR data, similar to approaches used by several teams in CXR-LT 2023. In contrast, Team \texttt{H} reported that their proposed model achieved superior performance without pretraining.

\paragraph{Ensemble Learning and Data Augmentation}

As with CXR-LT 2023, many top solutions (eight out of nine) employed a variety of ensemble learning strategies to improve generalization \citep{ganaie2022ensemble,fort2019deep}. Teams \texttt{B}, \texttt{C}, \texttt{F}, \texttt{G}, and \texttt{I} created ensembles across different model architectures; Teams \texttt{A} and \texttt{E} formed ensembles by using different image resolutions; and Team \texttt{H} constructed an ensemble strategy based on multiple views of the same image, but using the same model architecture across these views. In addition to ensemble learning, \textit{all} teams incorporated image augmentation, a well-established technique for enhancing generalization \citep{xu2023comprehensive}. Notably, Team \texttt{A} leveraged a diffusion model to generate synthetic images to augment rare tail classes.

\paragraph{Loss Re-weighting}

To address the long-tailed distribution of labels, five out of the nine top-performing solutions adopted loss re-weighting techniques to boost the importance of rare tail classes. These five teams (\texttt{B}, \texttt{C}, \texttt{D}, \texttt{F}, and \texttt{G}) all utilized a weighted asymmetric loss \citep{ridnik2021asymmetric}, which is specifically designed for handling imbalanced multi-label classification scenarios. Additionally, Team \texttt{G} implemented a weighted binary cross-entropy loss alongside this approach. The widespread adoption of the weighted asymmetric loss function can be attributed to its success in CXR-LT 2023, where it was employed by top-ranking teams. It is worth noting that two of the top solutions in Task 3 opted not to use weighted losses, primarily due to a lack of information about the distribution of the five unseen classes. However, one of the top solutions in Task 3 adopted a weighted loss approach by converting the zero-shot problem into a few-shot problem, leveraging prior knowledge about the unseen classes extracted from text-based descriptions.

\paragraph{Multimodal Vision-Language Learning}

Multimodal vision-language learning has recently gained popularity in deep learning for radiology, particularly as a pretraining approach using paired CXR images and free-text radiology reports \citep{chen2019deep,yan2022clinical,delbrouck2022vilmedic,moon2022multi,li2024llava,moor2023med}. This year, eight of the nine teams successfully leveraged both image and text data in some form. For example, Team \texttt{C} employed a combination of image-to-image and text-to-image contrastive learning to enhance feature representation. Meanwhile, Team \texttt{D} employed the ML-Decoder \citep{ridnik2023ml} classification head, which treats labels as text ``queries" that interact with image features via cross-attention. Notably, all three teams in Task 3 utilized multimodal vision-language models to enable zero-shot generalization to the five novel classes in Task 3.

\paragraph{ChatGPT/GPT-4}

With LLMs' increasing popularity in general and medical domains, three teams utilized ChatGPT or GPT-4 for Task 3. Team \texttt{G} used ChatGPT \citep{achiam2023gpt} to create fine-grained disease descriptions, potentially enhancing the performance of the text encoder. Team \texttt{H} leveraged GPT-4 to generate descriptive text augmentations for categorical labels, thereby facilitating zero-shot learning. In contrast, Team \texttt{I} opted not to use the provided labels and instead employed GPT-4 as an automatic labeler for MIMIC-CXR reports. They further incorporated additional CXR datasets, including some with bounding boxes, to support visual grounding supervision.

\paragraph{Implications of Synthetic Data for Long-Tailed Classification}

Team A's use of generative models to create synthetic data for rare classes appears to be a promising approach, as reflected in their performance. Synthetic data has the potential to mitigate extreme class imbalance by supplementing underrepresented classes, especially in domains where real data collection is costly or slow. However, it also raises questions about data fidelity, domain shift, and overfitting. As synthetic data generation techniques (e.g., diffusion models, GANs) continue to evolve, their role in addressing long-tailed medical image classification warrants further investigation, particularly regarding trustworthiness, generalizability, and clinical utility.
\subsection{Limitations and future work}


A key limitation of this study is the reliance on the MIMIC-CXR dataset, which was collected at a single academic medical center in the United States. As a result, the data may reflect institution-specific patient demographics, disease prevalence, imaging protocols, and equipment characteristics. These factors could limit the generalizability of the models to other geographic regions, healthcare systems, or clinical workflows. Although several top-performing teams incorporated additional publicly available CXR datasets (e.g., CheXpert, PadChest, VinDr-CXR) during training to enhance robustness, the final evaluation and leaderboard rankings were based solely on MIMIC-CXR test data. To more rigorously assess model generalization and ensure broader clinical applicability, future editions of the challenge should incorporate external test sets drawn from diverse institutions and populations. This would facilitate a more comprehensive evaluation of cross-site transferability and reveal potential sources of dataset shift or subgroup-specific bias.

Additionally, addressing bias in the models is critical. Existing studies have shown that deep neural networks trained on single-institution CXR datasets often exhibit disparities in predictive performance linked to factors like race and sex \citep{seyyed2020chexclusion, lin2023improving}. While \cite{seyyed2020chexclusion} observed that training on larger, multi-institutional datasets could help mitigate these disparities, their work focused on binary classification tasks. To date, no research has specifically examined bias in long-tailed, multi-label, and zero-shot classification tasks. Future investigations could explore methods to tackle these challenges, ensuring that models are both fair and generalizable across diverse populations and settings through rigorous subgroup analysis and multi-site validation.

Similar to most publicly available CXR benchmark datasets, the CXR-LT dataset is constrained by inherent label noise resulting from automatically extracted text-mined labels \citep{abdalla2023hurdles}. However, with the advancement of LLMs, recent studies \citep{wei2024enhancing} have demonstrated that GPT-4 can potentially generate more accurate labels for CXR datasets than traditional methods such as rule-based approaches. In our study, Table \ref{tab:comp-results} supports this observation, showing that GPT-4 produces higher-quality labels as evidenced by improved mAP. Other LLMs may also surpass traditional methods in label generation, presenting a promising avenue to reduce label noise in the future. Future iterations of CXR-LT may leverage LLM-based labeling pipelines to generate structured labels for arbitrarily large, long-tailed CXR datasets, which have proven successful in recent efforts \citep{zheng2024large}. Additionally, as more classes are included, the prompts for LLMs become longer, which may cause performance degradation by overwhelming the model and potentially leading to forgetting some classes. To mitigate this issue, reframing the labeling task as a natural language inference (NLI) problem and focusing the prompt on one class at a time can be an effective strategy. Moreover, incorporating techniques like chain-of-thought (CoT) prompting can further enhance performance by improving reasoning and response generation. Alternatively, a knowledge graph can be employed to separate the classes into different subgroups before applying LLM-based labeling, providing a structured and systematic approach to addressing this challenge.

Moreover, while it is challenging to obtain sufficient samples for deep learning training through manual annotation by radiologists due to the prohibitively high costs and time requirements \citep{zhou2021review}, providing a ``gold standard" dataset for testing purposes remains feasible. In this work, we leveraged and publicly released such a dataset with more reliable labels. However, this dataset was annotated by graduate students reviewing the clinical report text. In the future, this dataset could benefit from consensus re-annotation by radiology residents or attendings to enhance its quality. Additionally, manually annotating an external dataset for validation purposes could further enhance the evaluation of proposed methods, providing more reliable and accurate performance benchmarks.

As outlined in the overview of CXR-LT 2023 \citep{holste2024towards}, zero-shot classification can be the ideal approach for clinically viable long-tailed medical image recognition, enabling adaptation to any novel finding.
This year, the top three teams in Task 3 all utilized vision-language models to tackle this challenge, highlighting their potential. However, there remains significant room for improvement in several areas: aligning image and text representations more effectively, extracting information about unseen classes from textual data, and accurately detecting abnormal regions in images. Furthermore, efficient fine-tuning of vision-language models or instruction tuning will be crucial in addressing the challenges associated with the zero-shot disease classification problem.

Despite recent methodological advances, mean Average Precision (mAP) and F1 scores for chest X-ray (CXR) disease classification remain relatively modest. This raises a critical question: Are these performance metrics clinically acceptable? For example, an mAP in the range of 0.28--0.52 indicates performance well above random chance, yet likely falls short of the reliability required for autonomous clinical use. Several factors contribute to these limitations. First, extreme class imbalance -- especially with rare findings -- can skew performance and reduce sensitivity for less-represented diseases. Second, both training and evaluation datasets often contain label noise due to weak supervision or annotation inconsistencies. Third, chest X-ray as an imaging modality inherently lacks the resolution or contrast to clearly distinguish certain pathologies, particularly those with subtle or overlapping visual cues. Additionally, most models rely on thresholded probabilistic outputs, which can suffer from calibration issues, further affecting the reliability of decisions. To mitigate these challenges and improve real-world utility, future research may explore approaches such as multimodal decision fusion, calibrated confidence estimation, clinician-in-the-loop validation, and active learning techniques for better rare class sampling. Moreover, reporting per-class metrics and conducting failure mode analysis can help contextualize model performance, guiding more informed deployment strategies in clinical settings.

\section{Conclusion}

In summary, we organized CXR-LT 2024 to address the challenges of long-tailed, multi-label disease classification and zero-shot learning from chest X-rays. For this purpose, we have curated and released a large, long-tailed, multi-label CXR dataset containing 377,110 images, each labeled with one or more findings from a set of 45 disease categories. Additionally, we have provided a publicly available ``gold standard" subset with human-annotated consensus labels to facilitate further evaluation. Finally, we outline a pathway to enhance the reliability, generalizability, and practicality of methods, with the ultimate goal of making them applicable in real-world clinical settings.

\section*{Declaration of competing interest}

The authors declare the following financial interests/personal relationships which may be considered as potential competing interests: R.M.S has received royalties for patent or software licenses from iCAD, Philips, PingAn, ScanMed, Translation Holdings, and MGB, as well as research support from a CRADA with PingAn. The remaining authors declare that they have no known competing financial interests or personal relationships that could have appeared to influence the work reported in this paper.

\section*{Acknowledgments}

This work was supported by the National Library of Medicine [grant number R01LM014306], the NSF [grant numbers 2145640, IIS-2212176], the Amazon Research Award, Cornell–HKUST Global Strategic Collaboration Award, the Artificial Intelligence Journal, the National Institute of Health [grant number R01EB017205], DS-I Africa [grant number U54TW012043-01],  Bridge2AI [grant number OT2OD032701], and the National Science Foundation through ITEST \#2148451. It was also supported by the NIH Intramural Research Program, National Library of Medicine and Clinical Center.

\bibliographystyle{unsrtnat}
\bibliography{ref}

\newpage
\appendix
\setcounter{table}{0}
\setcounter{figure}{0}
\renewcommand\figurename{Supplementary Figure} 
\renewcommand\tablename{Supplementary Table}


\includepdf[pages=-]{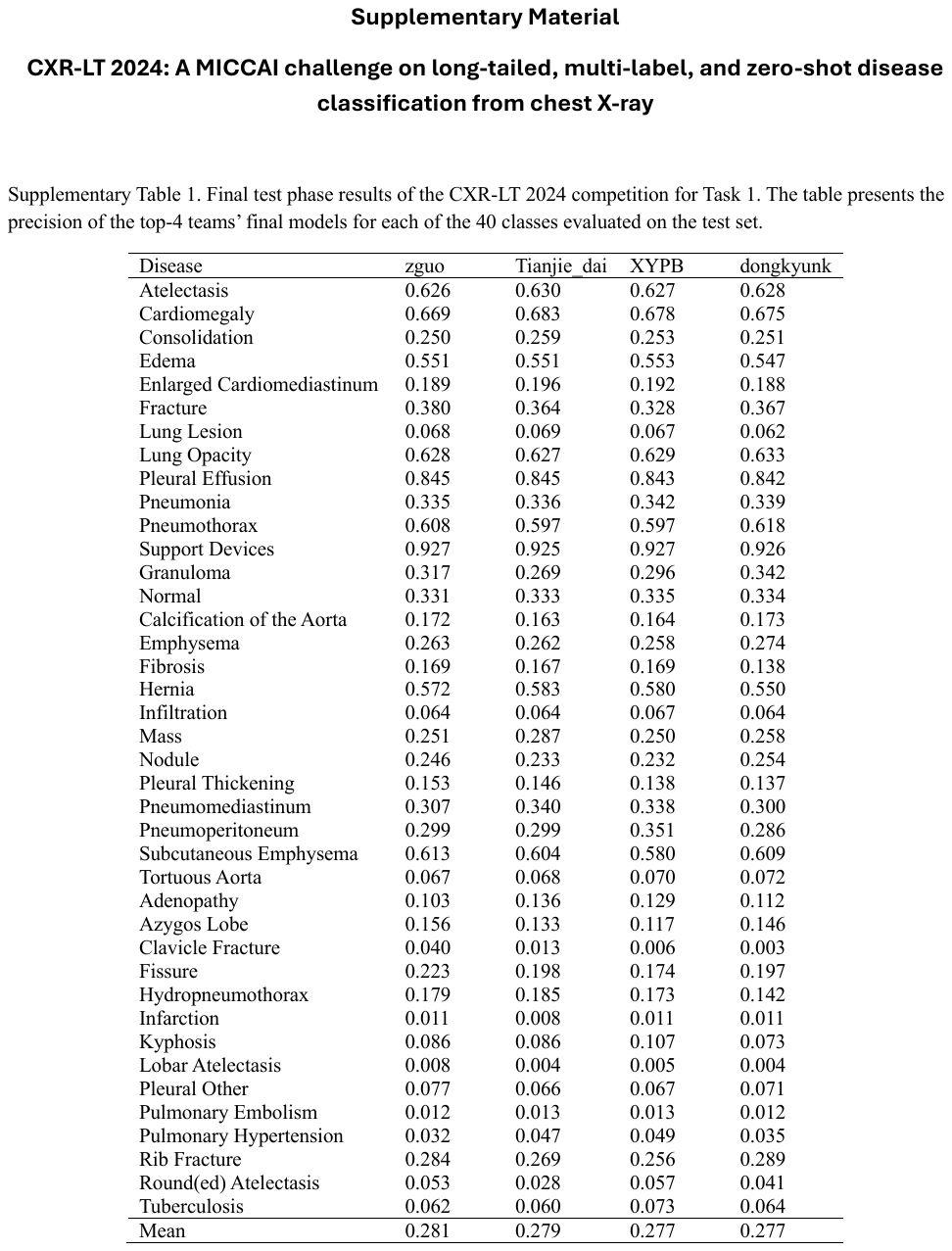}

\end{document}